%% file: arxiv.tex
\documentclass[runningheads]{llncs}

\usepackage[mobile]{eccv}

\usepackage{eccvabbrv}

\usepackage{graphicx}
\usepackage{booktabs}

\usepackage{comment}

\usepackage[accsupp]{axessibility}  

\usepackage[pagebackref,breaklinks,colorlinks,citecolor=eccvblue]{hyperref}

\usepackage{orcidlink}

\begin{document}

\title{Human Hair Reconstruction \\ with Strand-Aligned 3D Gaussians} 
\newcommand{\methodname}[0]{Gaussian Haircut}

\titlerunning{\methodname}

\author{Egor Zakharov\inst{1}\orcidlink{0000-0002-9880-9531} \and
Vanessa Sklyarova\inst{1,2}\orcidlink{0000-0002-8883-9972} \and
Michael Black\inst{2}\orcidlink{0000-0001-6077-4540} \and
Giljoo Nam\inst{3}\orcidlink{0000-0002-1822-1501} \and \\
Justus Thies\inst{2,4}\orcidlink{0000-0002-0056-9825} \and
Otmar Hilliges\inst{1}\orcidlink{0000-0002-5068-3474}
}

\authorrunning{E. Zakharov et al.}

\institute{ETH Zürich, Switzerland \and 
Max Planck Institute for Intelligent Systems, Tübingen, Germany \and 
Meta, Pittsburgh, USA \and
Technical University of Darmstadt, Germany}

\maketitle
\input{parts/notation}
\input{parts/abstract}
\input{parts/intro}
\input{parts/related}
\input{parts/method}
\input{parts/experiments}
\input{parts/conclusion}
\input{parts/acknowledgements}

%
%
\bibliographystyle{splncs04}
\bibliography{main}

\appendix

\input{parts_suppmat/method}
\input{parts_suppmat/experiments}

\end{document}

%% file: parts/notation.tex
\newcommand{\EZ}[1]{\iftrue {\color{orange}[EZ: #1]}\else {}\fi} 
\newcommand{\VS}[1]{\iftrue {\color{blue}[VS: #1]}\else {}\fi} 
\newcommand{\JT}[1]{\iftrue {\color{red}[JT: #1]}\else {}\fi} 
\newcommand{\MB}[1]{\iftrue {\color{green}[MB: #1]}\else {}\fi} 
\newcommand{\GN}[1]{\iftrue {\color{teal}[GN: #1]}\else {}\fi} 

%% file: parts/abstract.tex
\begin{abstract}
We introduce a new hair modeling method that uses a dual representation of classical hair strands and 3D Gaussians to produce accurate and realistic strand-based reconstructions from multi-view data.
In contrast to recent approaches that leverage unstructured Gaussians to model human avatars, our method reconstructs the hair using 3D polylines, or \emph{strands}.
This fundamental difference allows the use of the resulting hairstyles out-of-the-box in modern computer graphics engines for editing, rendering, and simulation.
Our 3D lifting method relies on unstructured Gaussians to generate multi-view ground truth data to supervise the fitting of hair strands.
The hairstyle itself is represented in the form of the so-called \emph{strand-aligned} 3D Gaussians.
This representation allows us to combine strand-based hair priors, which are essential for realistic modeling of the inner structure of hairstyles, with the differentiable rendering capabilities of 3D Gaussian Splatting.
Our method, named \emph{Gaussian Haircut}, is evaluated on synthetic and real scenes and demonstrates state-of-the-art performance in the task of strand-based hair reconstruction.
For more results, please refer to our project page: \url{https://eth-ait.github.io/GaussianHaircut}.
\keywords{3D Reconstruction \and Digital Humans \and Hair Modeling}
\end{abstract}

%% file: parts/intro.tex
\section{Introduction}
\label{sec:intro}

\input{figures/figure_main}

In recent years, human avatars have attained unprecedented levels of photorealism~\cite{qian2023gaussianavatars, Jiang2023HiFi4GHH, Saito2023RelightableGC, xu2023gaussianheadavatar, kirschstein2023nersemble, Wang2022NeuWigsAN, Lombardi2021MixtureOV, Wang2021HVHLA}.
This advancement has been largely driven by neural modeling and rendering methods that are incompatible with existing graphics pipelines.
However, many downstream applications require the resulting avatars to be deployable in physically simulated environments.
In particular, the de facto ``gold standard'' representation of human hair for simulation and rendering is based on strands, i.e., 3D curves.
Yet, strand-based reconstruction of hairstyles from real-world data, such as multi-view images, remains a challenging and heavily underconstrained problem, as a substantial part of hair geometry remains occluded in the image-based captures.

To address this, we leverage latent strand and hairstyle priors introduced by Rosu et al.~\cite{Rosu2022NeuralSL} and Sklyarova et al.~\cite{sklyarova2023neural_haircut} to produce strand-based reconstructions of hairstyles that have a realistic internal geometrical structure.
Following previous approaches, we first obtain a line-based reconstruction of the visible part of the hairstyle, where the 3D line segments are used to approximate the hair strands locally.
We then use this reconstruction to fit the strands with the aid of the pre-trained priors.
The classical approach for 3D hair modeling~\cite{Paris2004CaptureOH}  uses oriented gradient-based filters, the so-called orientation maps, to approximate projected line directions of the hair strands from the RGB images.
However, these maps are inherently noisy~\cite{Nam2019StrandAccurateMH, Rosu2022NeuralSL, sklyarova2023neural_haircut} and hard to use directly for strand fitting.
Thus, we use 3D Gaussians to perform an initial 3D lifting and de-noising of these orientation maps and use the former to supervise the reconstruction of hair strands.
To introduce extra details and fidelity into the reconstructions, we model a hairstyle using a dual representation of classical hair strands and 3D Gaussian primitives. 
The resulting method, called {\em Gaussian Haircut},  takes multi-view images as input and produces realistic 3D strands as output.

Our approach is inspired by recent work on 3D Gaussian Splatting (3DGS) \cite{kerbl3Dgaussians}, which is capable of reproducing high-fidelity details in novel-view renders of static scenes and objects.
Here, we utilize the 3D Gaussians in a novel way to enable realistic \emph{and} animatable hairstyle reconstruction.
We emphasize that our goal differs significantly from the original task of 3DGS, which reconstructs an \textit{unstructured set} of 3D Gaussian primitives incompatible with physics-based hair simulation.
These primitives cover only the visible hair surface, while we aim to reconstruct a hairstyle that includes the inner structure required to perform the simulation.
Both 3DGS and follow-up work~\cite{Saito2023RelightableGC} show that the learned Gaussian primitives are often well-aligned with thin structures in a scene, such as hair strands, without extra supervision.
We leverage this observation to compute a 3D orientation field of hair strands from the multi-view input images, which we refer to as 3D line lifting.
To achieve that, we render the directions of the highest variance of the Gaussian covariance matrices to obtain 2D orientation maps and compare them with ground-truth maps produced using Gabor filters.
Moreover, we introduce a camera refinement procedure into our Gaussian lifting approach designed to recover a more accurate 3D hair shape.

Having obtained accurate outer hair 3D orientations, we optimize a corresponding strand-based hairstyle using a strand-based hairstyle prior. 
We employ a dual representation to represent a hairstyle during our coarse-to-fine optimization consisting of classical 3D hair strand polylines and 3D Gaussians.
The coarse fitting relies on using a latent space to optimize the hair strands, while during fine fitting, we directly optimize the parameters of the decoded 3D polylines.
During the coarse-to-fine optimization, we employ differentiable rasterization of the hair strands to achieve a high accuracy in the reconstructed hairstyle.
Specifically, we use 3D Gaussians attached to the line segments of the strand polylines and propagate the gradients from their rasterization directly into the latter.
We supervise the reconstructions using both photometric and geometric constraints.
To supervise the geometry, we employ the 2D projections of the 3D orientation field obtained during the line lifting stage.
To help the reconstruction process, we also develop a hair upsampling technique that facilitates the effective use of the photometric constraints during the coarse optimization stage and a method for prior-based optimization of the strands in the domain of 3D polylines.
Our method is illustrated in Fig.~\ref{fig:main}.

\medskip

We evaluate our approach on both real and synthetic scenes and achieve improvements in both reconstruction speed and quality compared to the state-of-the-art.
Our contributions can be summarized as follows:
\begin{itemize}
    \item we propose a new \textbf{3D line lifting scheme} that uses a modified 3DGS reconstruction technique to lift 2D orientation maps into a 3D field while also providing refinement of the camera parameters;
    \item we introduce a \textbf{dual representation of hair strand polylines and 3D Gaussians} to achieve differentiable rasterization of hair strands and leverage photometric constraints for strand-based hair reconstruction;
    \item based on these components, we propose a \textbf{coarse-to-fine optimization method for prior-guided hair reconstruction} that leverages both latent and explicit representations of the hairstyle.
\end{itemize}

%% file: figures/figure_main.tex
\begin{figure}[!t]
    \begin{center}
    \includegraphics[width=\linewidth]{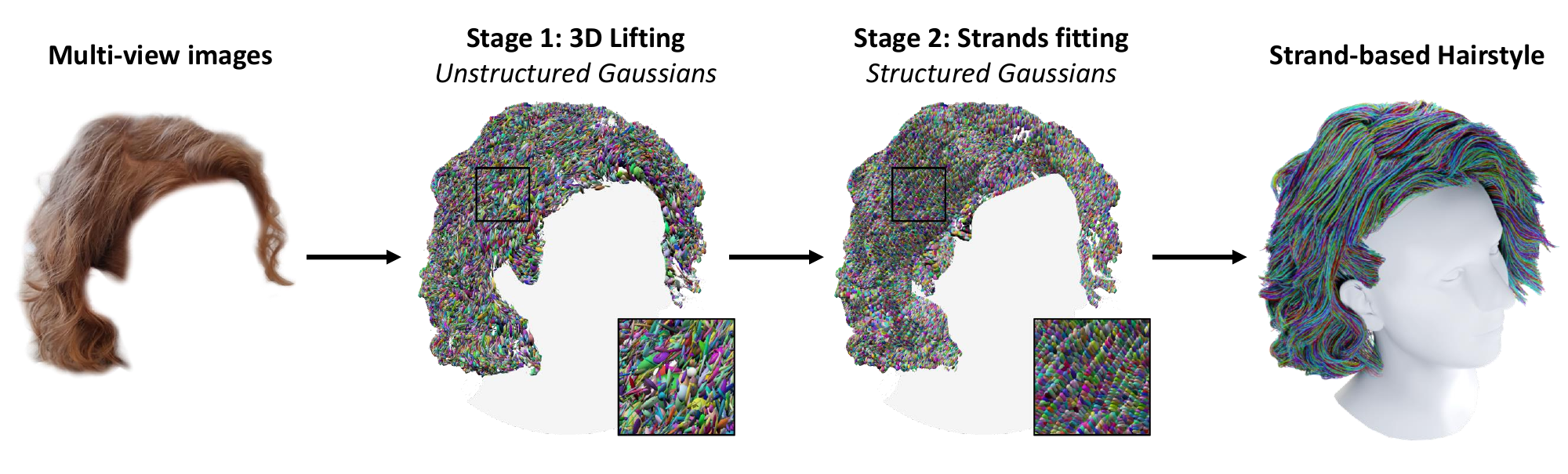}
    \caption{
    {\em Gaussian Haircut} works with multi-view images and uses strand-aligned 3D Gaussians to reconstruct a hairstyle.
    In the first stage, \emph{3D lifting}, we reconstruct the scene using unstructured primitives in the form of Gaussians.
    These unstructured primitives are then used in the second stage, \emph{strands fitting}, to supervise our dual hair strand representation consisting of 3D Gaussians that are attached to hair strands.
    As a result, we produce a realistic strand-based hairstyle that can be rendered, edited, and animated using classical computer graphics techniques.
    } 
    \label{fig:main}
    \end{center}
\end{figure}

%% file: parts/related.tex
\section{Related Work}
\label{sec:related}

\subsection{3D Gaussian Splatting for Head Avatars}
3D meshes are the most widely used representation for head avatars~\cite{FLAME:SiggraphAsia2017,cao2013facewarehouse}.
While a 3D mesh efficiently depicts surface-like areas such as faces, it struggles to represent complex geometries like hair accurately.
To overcome this limitation, volumetric representations like a mixture of volumetric primitives or neural radiance fields have been combined with 3D meshes to better portray hair and beards~\cite{lombardi2021mixture,garbin2022voltemorph,zielonka2023instant}.
However, volume rendering typically demands substantial computational resources.
Recently, there has been a shift towards using 3D Gaussian Splatting (3DGS) as a base representation for avatars~\cite{Saito2023RelightableGC,qian2023gaussianavatars}.
This approach achieves both photo-realistic appearance and real-time performance.
Despite these advancements, most studies~\cite{xiang2023flashavatar,xu2023gaussianheadavatar,chen2023monogaussianavatar,dhamo2023headgas,zhao2024psavatar,rivero2024rig3dgs} typically use 3D morphable face models such as FLAME~\cite{FLAME:SiggraphAsia2017} and do not model hair as a separate layer.
Instead, hair was treated as a mere component of the head, without any special consideration.
This restricts the potential applications of head avatars, particularly in hair motion and stylization. 

\subsection{Strand-based Hair Modeling}
Strands are the standard representation for high fidelity 3D hair modeling in  academia~\cite{Yuksel2009HairM,piuze2011generalized,shen2023CT2Hair} and industry~\cite{chiang2016practical,fascione2018path}.
Strands have several advantages over volumes or meshes, being particularly effective for physics-based simulations of hair~\cite{fei2017multi,daviet2023interactive,hsu2023sagfree}.
Strands also enable intuitive geometric editing by directly manipulating the length and curvature of strands~\cite{xing2019hairbrush,shen2020deepsketchhair,sklyarova2023haar,zhou2023groomgen}.
However, due to the geometric complexity of human hair and the large number of strands required, realistic hair modeling is labor-intensive  even for skilled 3D artists. 
To overcome this issue, image-based hair modeling methods have been developed to automate the 3D hair modeling process from captured photographs.
A common approach for image-based hair modeling involves the use of 2D orientation maps~\cite{Paris2004CaptureOH}.
Strand-based hair models are optimized to align with these 2D orientation maps when projected onto screen space~\cite{paris2008hair,luo2012multi,luo2013structure,luo2013wide,Paris2004CaptureOH,chai2015high,Chai2016AutoHairFA,zhang2017data,zhang2018modeling,Nam2019StrandAccurateMH}.
A notable limitation of image-based hair modeling is the lack of information about the interior parts of the hair, which are invariably occluded in images.
To overcome this challenge, Neural Haircut~\cite{sklyarova2023neural_haircut} employs a pretrained diffusion-based hairstyle prior to infer the interior geometry.
Despite these efforts, previous work represents hair as 3D polylines, which do not account for thickness in the geometry or appearance modeling.
These approaches typically have a separate stage to learn the appearance of 3D polylines from images~\cite{Rosu2022NeuralSL,sklyarova2023neural_haircut}.
In contrast, our proposed dual hair strand representation utilizes a sequence of anisotropic 3D Gaussians attached to a hair strand polyline, thereby offering enhanced expressiveness for both geometry and appearance modeling.

In the concurrent work of Luo et al.~\cite{Luo2024GaussianHairHM}, 3D Gaussians are also used to reconstruct human hair strands from multi-view images.
In contrast to this approach, which uses studio captures with uniform lighting, our method works with an unconstrained capture setup.
To enable this, we introduce a new camera optimization approach for 3D Gaussians.
Moreover, \cite{Luo2024GaussianHairHM} uses 3D constraints to optimize the hair strands, similar to Neural Haircut~\cite{sklyarova2023neural_haircut}, while our method fully relies on our new differentiable rendering scheme.

%% file: parts/method.tex
\section{Method}
\label{sec:method}

\input{figures/figure_aux}

Our method reconstructs 3D strand-based hairstyles from multi-view images.
We pre-process the data by calculating initial estimates for the cameras, segmentation masks, and image gradients, or the so-called \emph{orientation maps}.
The reconstruction is then performed in two stages, see Fig.~\ref{fig:main}.

For the first \emph{3D line lifting} stage, we use a combination of 3D Gaussian Splatting~\cite{kerbl3Dgaussians} and BARF~\cite{Lin2021BARFBN} to optimize the parameters of the training cameras and perform 3D lifting of the scene, including orientation maps.
The latter are directly embedded into the geometry of the resulting Gaussians.
Our main idea is to use a covariance matrix to represent the 3D orientation field of hair strands.
In our approach, the direction of maximum variance of each Gaussian corresponds to the direction of the hair strand, and the variance of the orthogonal directions gives us the natural measure of its uncertainty, which we account for during the rendering process.
As a result, we obtain a scene representation that produces highly realistic renders for both color and orientation maps, as the resulting Gaussians are aligned with the hair strands.
However, these primitives are \emph{unstructured}, as they only lie on the outer hair surface, see Fig.~\ref{fig:aux}~(a).
We thus use them to generate the multi-view renders and reconstruct the geometry of the hair strands during the second stage using \emph{structured} primitives.

In the \emph{hair strands fitting} stage, we optimize the strand-based hairstyle in a coarse-to-fine manner.
For the coarse optimization step, we parameterize hair as a latent texture map~\cite{Rosu2022NeuralSL}.
We use the Gaussians obtained during the first stage to create ground truth for photometric and geometric losses.
After the coarse optimization, we decode the strands into an explicit hair map and optimize it directly in a fine-grained fitting step.
During both of these steps, we follow~\cite{sklyarova2023neural_haircut} and regularize the hairstyle using a pre-trained hairstyle diffusion model~\cite{Karras2022ElucidatingTD} to increase the realism of its internal structure.
We also utilize the 3D Gaussian Splatting framework to incorporate differentiable rendering of the hair strands.
We attach a Gaussian to each line segment on the strand to achieve that.
The scale of these primitives has only one degree of freedom, which is proportional to the length of the line segment, while the other two are fixed to a small pre-defined value.
This allows us to introduce fine-grained details into the visible part of the hair reconstruction.

\subsection{3D Line Lifting with Unstructured Gaussians}

We rely on the 3D Gaussian Splatting approach to perform an initial scene reconstruction.
We modify it to incorporate camera optimization, as off-the-shelf SfM~\cite{schoenberger2016sfm} methods do not achieve high accuracy when localizing the cameras in the scenes centered around the hair~\cite{sklyarova2023neural_haircut}.
Therefore, we employ a learnable 6-DoF camera parameterization from BARF~\cite{Lin2021BARFBN} as a residual to the initial estimation produced by SfM and train it alongside the 3D Gaussians using gradient-based optimization.

Following~\cite{kerbl3Dgaussians}, each reconstructed primitive is parameterized by a learnable mean $\mu$, scaling coefficients $s$, rotation quaternion $q$, and opacity $o$. 
Their covariance matrix can then be expressed as $\Sigma = R S S^T R^T$, where $R$ is a matrix form of the quaternion $q$, and $S = \text{diag}(s)$.
Each Gaussian also has a set of learnable features, which include the spherical harmonic coefficients $f$ for the view-dependent color, a hair segmentation label $l$, and confidence value $\tau$ for the 3D orientation of the Gaussian.
The rendering proceeds by first projecting each Gaussian into the screen space, resulting in a mean $\mu'$ and a covariance matrix $\Sigma'$.
The Gaussians are then sorted according to their depth, and for each pixel $p$, their features $c_i$ are rendered using $\alpha$-blending: 
\begin{equation}
    C_p = \sum_{i=1}^{N} T^i_p \alpha^i_p c_i,\quad T^i_p = \prod_{j=1}^{i-1} (1 - \alpha^j_p),\quad T_p^1 = 1,
    \label{eq:volume_rendering}
\end{equation}
\begin{equation}
    \alpha^i_p = o_i \exp \Big( -\frac{1}{2} (p - \mu'_i)^T \Sigma'_i (p - \mu'_i) \Big) .
    \label{eq:opacity_formula}
\end{equation}
As a result, we obtain rendered color, which is produced via the original Gaussian splatting approach, hair segmentation label $l_p$, orientation confidence value $\tau_p$, and a rendered silhouette of the Gaussians denoted as $s_p$:
\begin{equation}
    l_p = \sum_{i=1}^N T_p^i \alpha_p^i l_i,\quad \tau_p = \sum_{i=1}^N T_p^i \alpha_p^i \tau_i,\quad s_p = \sum_{i=1}^N T_p^i \alpha_p^i.
\end{equation}

In our approach, we use the covariance matrices to describe the local geometry of the hair strands in the 3D volume. To do that, we align the Gaussians with the hair strands via 3D lifting of the orientation maps.
We represent the primary 3D orientation as the direction of the highest variance and use the variance of the orthogonal directions as a measure of uncertainty.
Specifically, we obtain a rendered strand direction in pixel $p$, denoted as $\beta_p$, as follows:
\begin{equation}
    \beta_p = \sum_{i=1}^N T_p^i \alpha_p^i \beta_i,
\end{equation}
where $\alpha_p^i$ and $T_p^i$ depend on the full covariance matrix, as per Eq.~\ref{eq:volume_rendering}-\ref{eq:opacity_formula}, and $\beta_i$ denotes a direction of the highest variance of the Gaussian.

The Gaussians are trained using a gradient-based optimization procedure from the original work~\cite{kerbl3Dgaussians}. Our photometric constraints include $\mathcal{L}_\text{rgb}$, which consists of L1 and SSIM losses, $\mathcal{L}_\text{seg}$ that consists of an L1 loss that matches the rendered silhouette $s_p$ and hair segmentation $l_p$ to ground truth masks, and $\mathcal{L}_\text{dir}$ that supervises hair orientations. For the latter, we build upon an original formulation~\cite{Paris2004CaptureOH} and introduce a rendered confidence factor $\tau_p$ into the objective:
\begin{equation}
    \mathcal{L}_\text{dir} = \sum_p \tau_p \min \{ \text{d} (\beta_p, \hat\beta_p), \text{d} (\beta_p, \hat\beta_p) \pm \pi \} - \log \tau_p,
\end{equation}
where $\text{d}$ denotes the absolute angular difference between the directions, and $\hat\beta_p$ denotes a ground-truth direction in the pixel $p$.
Our resulting training objective for the Gaussian-based hair reconstruction stage can be written as follows:
\begin{equation}
    \mathcal{L}_\text{gaussian} = \mathcal{L}_\text{rgb} + \lambda_\text{seg} \mathcal{L}_\text{seg} + \lambda_\text{dir} \mathcal{L}_\text{dir}. 
    \label{eq:loss_stage1}
\end{equation}

\subsection{3D Hair Strands Reconstruction}

After the first stage, we obtain Gaussians matching the hairstyle's visible structure, including the geometry and strand orientations, and the optimized camera parameters.
This information is used for the 3D hair strands reconstruction.

The strand-based hairstyle is represented as a hair map $H$ corresponding to a scalp region of the 3D head model~\cite{FLAME:SiggraphAsia2017}.
This head model is fitted into the scene using a multiview optimization approach based on facial keypoints~\cite{sklyarova2023neural_haircut}.
Each texel of this map stores a 3D hair strand as a polyline: $S^k = \{ p_l^k \}$.
Since hair maps have an increasingly high number of degrees of freedom, we regularize the optimization process~\cite{Rosu2022NeuralSL} and introduce a latent hair map $Z$, which can be converted to and from $H$ using pre-trained strand decoder $\mathcal{G}$ and encoder $\mathcal{E}$:
\begin{equation}
    H = \mathcal{G}(Z),\quad Z = \mathcal{E}(H).
\end{equation}
During the coarse fitting step, we optimize the hair strands as a latent map $Z$, while for the fine step, we optimize an explicit hair map $H$, see Fig.~\ref{fig:aux}~(c).

Since latent hair map decoding operation is computationally expensive, in the coarse optimization step, this mapping can only be performed for a set of \emph{guiding} strands per training batch, which we will denote as $H'$, see Fig.~\ref{fig:aux}~(b).
Note, however, that the latent map $Z$ can be decoded into an arbitrarily large number of hair strands for the fine-fitting step.
The hair strands are optimized using two constraints: photometric losses, for which we differentiably rasterize the hair strands 
and latent diffusion-based regularization that follows~\cite{sklyarova2023neural_haircut} and ensures the realism of the internal part of the hairstyle.
Importantly, we use the ground truth generated using the Gaussian-based reconstruction from the previous stage, substantially reducing the noise in this objective, especially for the orientation maps.

\paragraph{Photometric losses.}
During differentiable rasterization, we assign Gaussians to each line segment $\{ p_l^k, p_{l+1}^k \}$ of the strand.
The generated primitive has the first element of its scaling vector $s_l^k$ proportional to the length of the strand, and the other two fixed to a small value, i.e.~$s_l^k = \{ 1 \slash 2 \, \cdot \, \| p_{l+1}^k - p_l^k \|_2 \, , \, \epsilon \,, \, \epsilon \}$.
Additionally, its rotation quaternion is set to align the $x$-axis with the segment direction of the hair strand.
We make all these Gaussians opaque, meaning $o_l^k = 1$, have unit orientation confidence $\tau_l^k$, and assign each one a trainable set of spherical harmonic coefficients $f_l^k$ responsible for rendering the color.
During the coarse fitting, these coefficients are predicted from the latent hair map using an appearance decoder $\mathcal{G}_\text{a}$, which copies the architecture of a strand decoder $\mathcal{G}$.
However, while the strand decoder $\mathcal{G}$ is pre-trained using a collection of synthetic hair strands and is frozen during the reconstruction process, following~\cite{Rosu2022NeuralSL}, the appearance decoder $\mathcal{G}_\text{a}$ is optimized from scratch for each scene.
Thus, the trainable parameters during the coarse optimization step include $Z$ and $\mathcal{G}_\text{a}$, while for the fine step, we directly optimize $H$ with associated coefficients $f_l^k$.

The rendering loss follows the Gaussians training objective $\mathcal{L}_\text{gaussian}$ from the first stage (Eq.~\ref{eq:loss_stage1}) and includes a color loss $\mathcal{L}_\text{rgb}$, a segmentation loss $\mathcal{L}_\text{seg}$, and an orientations loss $\mathcal{L}_\text{dir}$.
Instead of using the covariance matrices to extract the orientations $\beta_i$, we use the direction vectors $v_l^k = p_{l+1}^k - p_l^k$.
We can do that since, at this stage, we have access to the exact strand growth \emph{directions} instead of undirected lines.
However, during the coarse fitting stage, we can only decode a small set of guiding hair strands $H'$ per training batch from the latent map $Z$ because of memory constraints.
Thus, the rendered geometry may have holes, making photometric training losses ineffective.
To address this problem, we interpolate the guiding strands to produce a dense hair map $\hat{H}$, which we then use for rasterization.
The strand interpolation procedure follows~\cite{sklyarova2023haar} but is carried out in the domain of 3D coordinates instead of latent hair maps.
Note that we do not need such interpolation during the fine optimization stage, as at each training step we have direct access to the dense hairstyle.

\paragraph{Diffusion-based regularization.}
We follow~\cite{sklyarova2023neural_haircut} and employ a diffusion-based score distillation sampling~\cite{Poole2022DreamFusionTU} loss $\mathcal{L}_\text{sds}$ to regularize the hair geometry.
It is applied in the domain of the latent hair maps and uses a pre-trained latent diffusion for guide strand generation~\cite{sklyarova2023neural_haircut}.
While it can be incorporated seamlessly into our coarse optimization step by simply applying it to a subsampled version $Z'$ of the latent hair map $Z$, we also want to keep this regularization during the fine-grained strands optimization.
To achieve that, at each step, we encode a random subset of the hair strands into their latent representations using a pre-trained encoder $\mathcal{E}$ and interpolate them into texels with the same resolution as $Z'$.
We then apply a diffusion-based penalty to this low-resolution latent hair map, similar to the coarse stage.
For more details, please refer to the supplementary materials.

\paragraph{The final objective} for the hair strands fitting can be written as follows:
\begin{equation}
    \mathcal{L}_\text{strand} = \mathcal{L}_\text{rgb} + \lambda_\text{seg} \mathcal{L}_\text{seg} + \lambda_\text{dir} \mathcal{L}_\text{dir} + \lambda_\text{sds} \mathcal{L}_\text{sds}.
\end{equation}
As a result, we obtain a strand-based hair map that can be rendered statically using optimized spherical harmonics coefficients or dynamically via classical computer graphics engines~\cite{Blender}.

%% file: figures/figure_aux.tex
\begin{figure}[!t]
    \begin{center}
    \includegraphics[width=\linewidth]{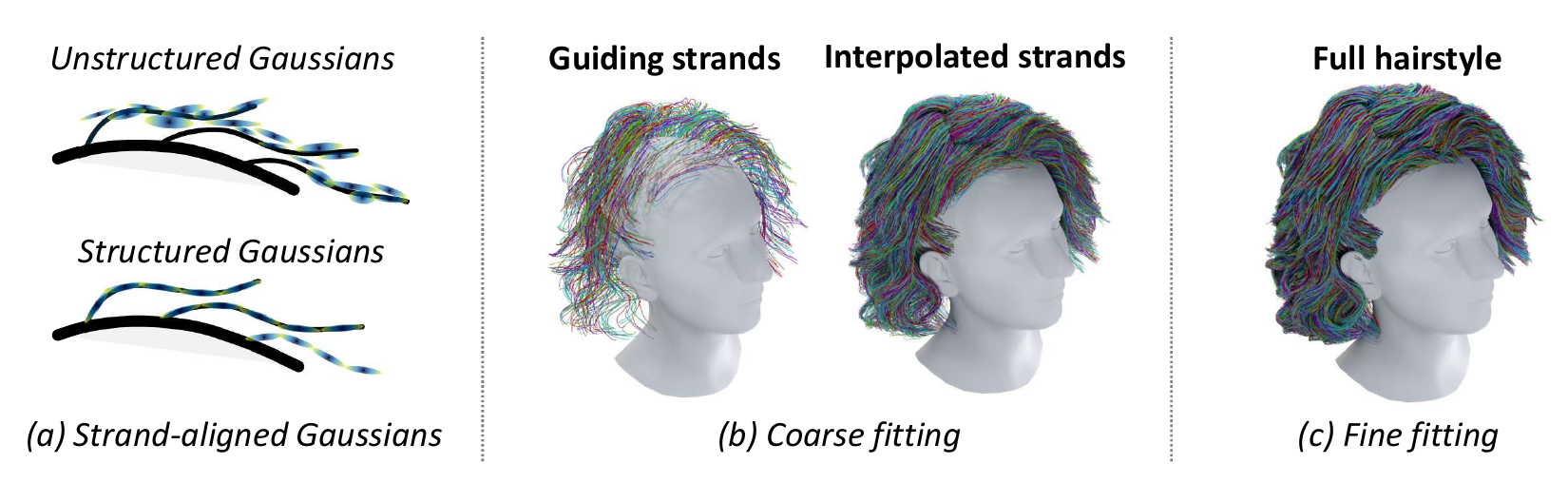}
    \caption{
         (a) In our work, we utilize both structured and unstructured sets of Gaussians. The former are attached to the strands and thus cover an entire hair volume, while the latter concentrate only on the visible part of the hair surface. (b) During coarse strand-based fitting, we decode only a set of guiding strands from the latent hair map at each training step, as generating a full hairstyle is computationally- and memory-expensive. We then convert these strands into a dense hair map suitable for rendering by conducting interpolation in the space of their 3D coordinates. (c) Finally, we conduct a fine strand-based optimization step. We decode a dense hairstyle from a latent map and directly optimize their coordinates instead of latent representations.
    } 
    \label{fig:aux}
    \end{center}
\end{figure}

%% file: parts/experiments.tex
\section{Experiments}

\subsection{Implementation Details}

We preprocess real-world data using COLMAP~\cite{schoenberger2016sfm} to produce the initial point cloud and camera estimates.
We use the default optimization hyperparameters from the base 3DGS approach~\cite{kerbl3Dgaussians} to train the Gaussians, which results in 30,000 optimization steps during the 3D lifting stage.
The camera parameters are optimized for the first 15,000 steps and are frozen afterward.
Their learning rate schedule follows the learning rates of Gaussian rotations and means.

During the strand fitting stage, we use pre-trained strand and hairstyle priors from~\cite{sklyarova2023neural_haircut}.
We use 1,000 guiding strands in the hair map $H'$ during coarse optimization, which are interpolated to 10,000 at each training step before rasterization.
For the fine fitting of the hairstyle, we decode a hair map $H$ containing 30,000 strands.
To generate a guiding latent hair map $Z'$ that is used for regularization, we randomly select 1,000 strands at each training step.
We encode them into the latent vectors and interpolate them into a latent map before calculating a score distillation sampling loss.
We use 20,000 and 10,000 optimization steps for the coarse and fine fitting, respectively.
The weights of the losses during both stages are set to $\lambda_\text{seg} = \lambda_\text{dir} = 10^{-1}, \lambda_\text{sds} = 10^{-2}$.
Reconstructing a hairstyle using our method takes up to six hours on a single RTX 4090.
For more details on implementation and training, please refer to the supplementary materials.

\subsection{Real-world Evaluation}

We evaluate our approach using monocular video sequences captured under unconstrained lighting conditions.

\paragraph{Baseline.} 
The main baseline we compare against is Neural Haircut~\cite{sklyarova2023neural_haircut}.
This method reconstructs strand-based hair geometry in the same scenario as ours, i.e.~using real-world multi-view image/monocular video data.
It consists of two stages.
During the first stage, it reconstructs coarse hair geometry using a layered neural signed distance function~\cite{Wang2021NeuSLN}, where the layers correspond to the hair and bust geometries.
The hair surface geometry also incorporates a 3D-lifted orientation field. 
During the second stage, the strands are parameterized as latent hair maps and optimized using the Chamfer distance between the strands and the visible part of the hair surface.
The authors also propose a differentiable hair rendering pipeline that relies on mesh-based rasterization to propagate photometric information into geometry.
However, this loss does not propagate high-frequency details and only slightly improves strand geometry.

\input{figures/comparison/comparison}

\paragraph{Comparison.} In contrast to this method, we primarily rely on differentiable rendering of the hair strands to learn their geometry.
Toward this end, we use a two-stage training approach that first does bundle adjustment and 3D lifting of orientation maps using unstructured Gaussians and then utilizes structured Gaussians to learn the hairstyle's shape via differentiable rasterization.
Our approach achieves considerable quality improvements in the resulting hairstyles, as seen in the qualitative evaluation, Fig.~\ref{fig:geom_compare}.
Another advantage of our method is the faster reconstruction timings.
We achieve more than ten times speed-up compared to the base approach with improved reconstruction quality.

\input{figures/rendering}

\input{figures/simulations/simulations}

\paragraph{Applications.}
The learned structured Gaussians can be used out-of-the-box for fast and photorealistic rendering of the reconstructed hairstyles.
We showcase the rendering results in Fig.~\ref{fig:rend_eval}.
These images can be generated in seconds, on par with the rendering time required to achieve similar visual quality with the classical rendering approaches, for example, using~\cite{unrealengine}.
They also have the upside of matching the lighting and the subject's hair color without requiring manual adjustment.
Also, due to the dual nature of our reconstructions, we can leverage the explicit geometry of the reconstructed strands to conduct a generalizable physics simulation driven by the motion of the head.
We showcase the simulation results in Fig.~\ref{fig:simulations} and the supplementary materials.
Note that achieving such high plausibility of simulated dynamics is only possible because our reconstructions are structured into strands, are connected to the parametric head mesh, and have realistic internal structures due to the use of hairstyle priors.

\subsection{Ablation Study}

\input{figures/ablation/ablation}

We ablate the importance of our method's components on real and synthetic scenes.
Following~\cite{sklyarova2023neural_haircut}, for synthetic evaluation, we use two hair models of straight and curly hairstyles~\cite{Yuksel2009HairM} that are widely used in the literature as benchmarks.

\paragraph{Orientation maps lifting.} First, we evaluate the efficiency of our 3D line lifting approach in producing the denoised orientation maps.
A standard hair orientation map calculation approach relies on Gabor filter banks~\cite{Paris2004CaptureOH}.
Thus, we can utilize a rendering engine~\cite{Blender} to produce realistic renders of synthetic hairstyles, which can be used to run orientation map extraction via these filters.
For quantitative evaluation, we also generate the ground-truth orientation maps directly from the underlying geometry using classical line rendering algorithm~\cite{woo1999opengl}.
Our method achieves the lowest average angular error of re-projections of $7^\circ$, followed by the error of  $8^\circ$ achieved by Gabor filters. 
These metrics are the average of 70 test views for two synthetic hairstyles.
Thus, our 3D line lifting gives better estimates for hair orientation maps than Gabor filters.
Moreover, we evaluate the effect of these synthesized orientation maps on the reconstructed hair geometry in real-world examples.
The results are in Fig.~\ref{fig:ablation_compare}, third column.
Note that with synthesized orientation maps, our method can model the geometry more accurately in regions with poor illumination and highly entangled hair strands, such as the one shown in the first row. 
Please refer to the supplementary materials for more examples.

\paragraph{Coarse-to-fine strands fitting.}
Next, we evaluate the effectiveness of our strand-fitting approach.
First, we showcase the results without the fine fitting step, Fig.~\ref{fig:ablation_compare}, second column, and without synthetic renders of orientation maps, third column.
We observe that our method fails to achieve high levels of accuracy in the resulting hairstyles with only the coarse optimization step or without using our proposed 3D orientation maps lifting methods, especially in the areas with a high density of strands and poor illumination: see the first row.
Then, we evaluate our proposed upsampling technique for hair rasterization, column fourth.
We observe that without using strand upsampling during the coarse stage, the fine fitting cannot converge to an accurate hairstyle; see the first, third, and fourth rows.
Finally, we evaluate the effectiveness of using orientation loss $\mathcal{L}_\text{dir}$, see column fifth.
The extended version of the ablation study with additional experiments is available in the supplementary materials.
There, we also validate the diffusion-based prior's efficiency.

\label{sec:experiments}

%% file: figures/comparison/comparison.tex
\begin{figure*}
    \includegraphics[width=\linewidth]{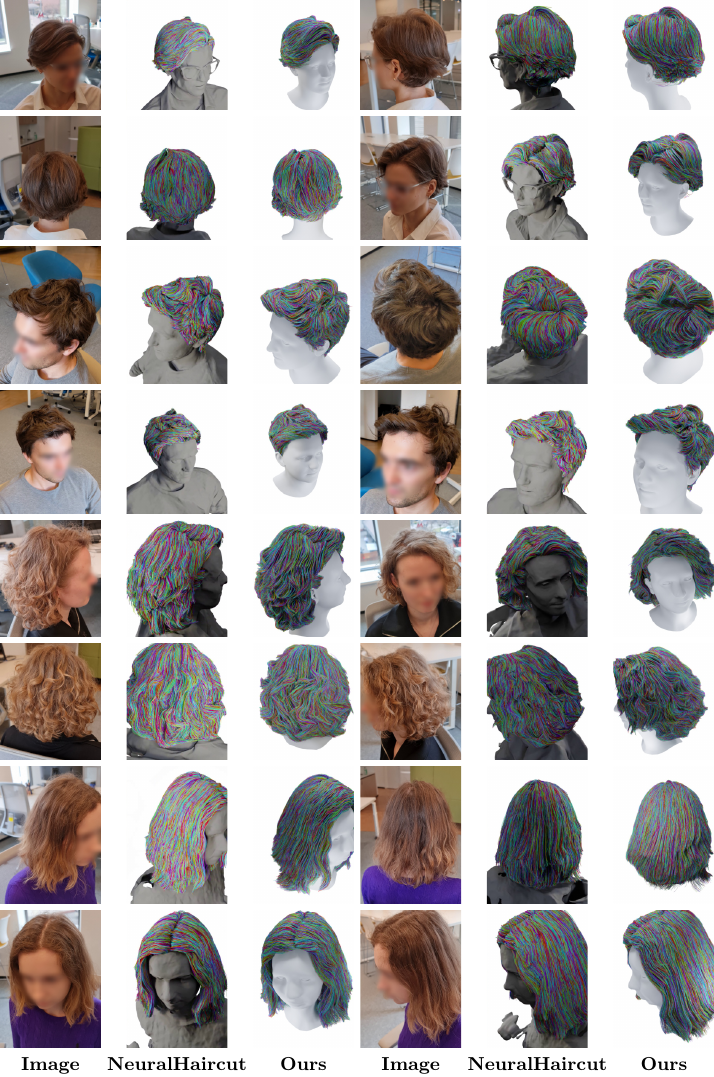}
    \caption{A qualitative comparison of the reconstructed strand-based geometry against Neural Haircut~\cite{sklyarova2023neural_haircut}. Our method has higher accuracy of the reconstructed hairstyles while achieving more than ten times improvement in the optimization speed.}
    \label{fig:geom_compare}
\end{figure*}

%% file: figures/rendering.tex
\begin{figure*}[!t]
    \includegraphics[width=\linewidth]{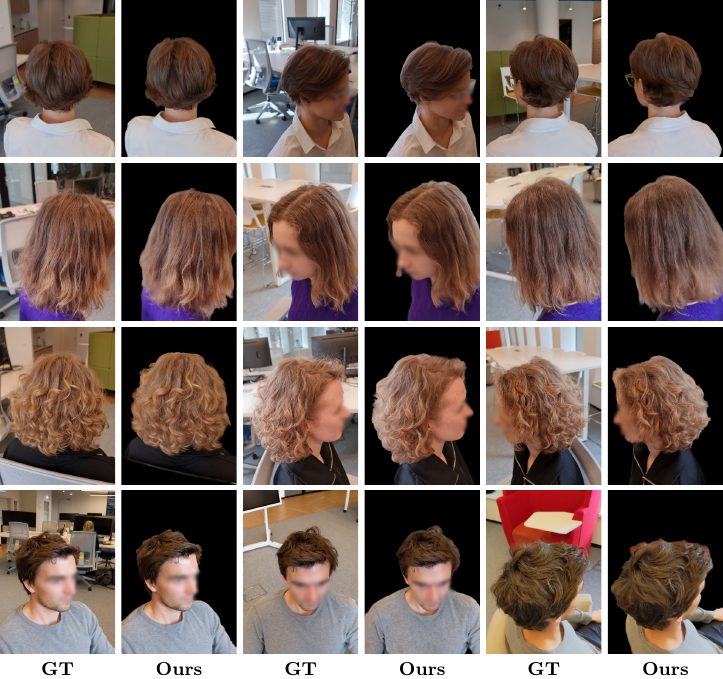}
    \caption{Our method can achieve photorealistic renders of strand-based geometry using structured Gaussians. We showcase the rendering results produced by our Gaussian-based strand rasterization method from test views. Since these images were not included in the training set and thus were not part of the bundle adjustment, there is a slight misalignment between the predicted images and the ground truth.}
    \label{fig:rend_eval}
\end{figure*}

%% file: figures/simulations/simulations.tex
\begin{figure*}[!t]
    \includegraphics[width=\linewidth]{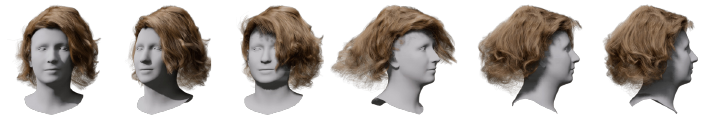}
    \caption{Hairstyles produced by our method can be easily imported into a computer graphics engine for editing, rendering, and simulation. Here, we show the simulation results of the reconstructed hairstyle in Unreal Engine~\cite{unrealengine}.}
    \label{fig:simulations}
\end{figure*}

%% file: figures/ablation/ablation.tex
\begin{figure*}[!t]
    \includegraphics[width=\linewidth]{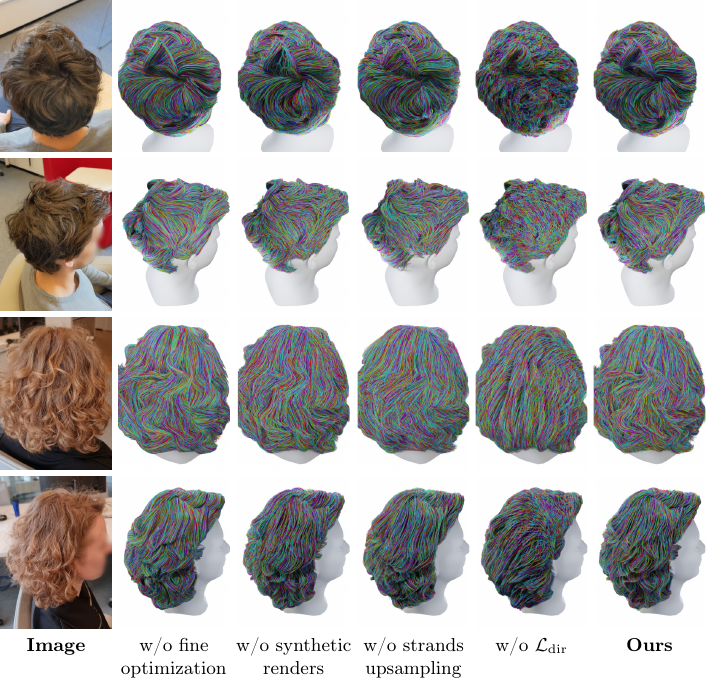}
    \caption{Ablation of the strands fitting stage. We show reconstructions obtained with only coarse stage fitting: {w/o fine optimization}. Then, we ablate the use of the orientation maps renders generated with unstructured Gaussians for geometry supervision: {w/o synthetic renders}; using upsampling of guiding strands during coarse strands fitting: {w/o strands upsampling}; not using orientation loss: {w/o $\mathcal{L}_\text{dir}$}. We observe that all our proposed components contribute to the final quality of the results.}
    \label{fig:ablation_compare}
\end{figure*}

%% file: parts/conclusion.tex
\section{Conclusion}
\label{sec:conclusion}
We have presented a method for accurate strand-based hair reconstruction.
To achieve that, we propose a new hair modeling method that consists of Gaussian-based 3D line lifting and a strand optimization approach that relies on our new dual representation called strand-aligned Gaussians.
We also introduce a coarse-to-fine optimization method for hair strands that achieves state-of-the-art performance in multi-view hair reconstruction using 360$^\circ$ head capture with unconstrained lighting conditions.
Our findings are validated via an extensive evaluation using real-world and synthetic data.
The limitations of our method are in many aspects similar to the base Neural Haircut approach.
For example, the strand-based hair prior struggles to model curly hairstyles, as seen in one of the examples in Fig.~\ref{fig:geom_compare}.
Also, our method is unsuited for reconstructing hairstyles with complicated internal structures, such as braids.
Thus, introducing support for more sophisticated hair geometries remains future work.
Our method also achieves a substantial speed-up compared to the baseline approaches, outperforming the optimization speed of the previous state-of-the-art by more than ten times.
We also showcase how the hairstyles produced by our method can be directly imported into computer graphics engines, such as Unreal Engine, for use in physically rendered and simulated environments.
We believe that fast and accurate reconstruction of hairstyles from real-world data can open the way to exciting applications in telepresence, special effects, and gaming industries.

%% file: parts/acknowledgements.tex
\section*{Acknowledgements}

We sincerely thank Joachim~Tesch for his help with realistic hair simulations.
The data presented in the work was provided by the MPI-IS Capture Team and the Samsung AI Center.
We thank the MPI-IS Capture Team, especially Asuka~Bertler, Tsvetelina~Alexiadis, and Claudia~Gallatz, for their help and assistance in capturing data.
Egor~Zakharov was funded by the ``AI-PERCEIVE'' 2021 ERC Consolidator Grant. 
Vanessa~Sklyarova was supported by the Max Planck ETH Center for Learning Systems. 
Michael~Black conflict of interest disclosure: \url{https://files.is.tue.mpg.de/black/CoI_ECCV_2024.txt}.

%% file: parts_suppmat/method.tex
\section{Implementation Details}
\label{sec:impl_suppmat}
Below we describe each processing step of our proposed method in more detail.

\subsection{Preprocessing}
We experiment with video-based captures but can employ a similar pipeline to process image-based captures.
First, we extract the frames from the video with the framerate of 3 FPS.
To reduce the motion blur, among the frames spanning each $1/3$ of a second, we pick the one with the highest image quality score calculated via a HyperIQA~\cite{IQA} network.
We run COLMAP SfM~\cite{schoenberger2016sfm} algorithm for the extracted frames to estimate the initial point cloud and the camera parameters.
To obtain the segmentation masks, we rely on Matte-Anything~\cite{matte_anything} prompt-based image matting system.
To produce the corresponding segmentation masks, we use the following prompts: ``hair'', ``face'', and ``human''.
We then filter the cases where segmentation masking fails.
For that, we have designed a heuristic in which we calculate the intersection between the hair and face and prune the frame if it is higher than $10\%$ of the body mask area.
Lastly, we calculate image quality scores again for the masked hair crops.
We split the filtered images into consecutive $128$ bins and picked the image with the highest score from each bin for training, resulting in $128$ training views.
These pre-processing steps closely follow the ones used in Neural Haircut~\cite{sklyarova2023neural_haircut}.
We rely on the pipeline proposed in~\cite{Paris2004CaptureOH} for orientation map calculation and use the code of Neural Haircut~\cite{sklyarova2023neural_haircut}.

\subsection{Guiding Strands Upsampling}
We follow HAAR~\cite{sklyarova2023haar} in the procedure that we use for the interpolation of the hair strands.
In this work, they perform upsampling using a regular grid of latent maps and blend the strands produced using nearest and bilinear interpolation methods.
In our case, we use irregular grids and thus use K nearest neighbors for both, replacing bilinear interpolation with the one that blends four nearest neighbors.
The KNN-based interpolation weights are inversely proportional to the distance between the interpolated strands' origins and the query origin in the space of texture coordinates.
Another change compared to HAAR is that we conduct the interpolation in the space of hair maps instead of latent maps.
Thus, we directly interpolate the 3D coordinates of the strands, each being defined within its own TBN (tangent-bitangent-normal) basis, which is defined using the scalp face where the strand originates~\cite{Rosu2022NeuralSL}.

\subsection{Diffusion-Based Prior}
Our diffusion-based prior follows Neural Haircut~\cite{sklyarova2023neural_haircut}.
We rely on Score Distillation Sampling (SDS)~\cite{Poole2022DreamFusionTU} to enforce this prior at the level of latent hair maps.
Thus, during the coarse strand fitting stage, we directly follow Neural Haircut~\cite{sklyarova2023neural_haircut} in applying it.
To utilize diffusion-based prior at the level of strand-based hair maps $H$, we use a pre-trained encoder $\mathcal{E}$ that maps a hair map $H$ into its latent version $Z$.
Specifically, we randomly sample a set of 1,000 guiding strands and map them into a regular grid of origins using an interpolation method described above.
As a result, we produce a guiding hair map $H'$, which we can encode into a guiding latent hair map $Z'$.
This map is then directly fed into a diffusion-based prior in order to calculate an SDS loss.

\subsection{Postprocessing}
After training, to achieve simulatable hairstyles, we post-process the resulting strands to ensure they have no intersections with the FLAME mesh.
To achieve that, we calculate a signed distance for all the points on the strands towards the head mesh.
Then, for the strands that intersect the mesh, we prune the intersection regions and then connect the beginning of each of the consecutive strand segment to the nearest vertex on the scalp.

%% file: parts_suppmat/experiments.tex
\section{Experiments}

\input{figures_suppmat/comparison/comparison}

\input{figures_suppmat/comparison_2/comparison_new}

\input{figures_suppmat/comparison_2/comparison2_2}

\input{figures_suppmat/short_hair}

\input{figures_suppmat/hairstep_comparison}

\input{figures_suppmat/limitations/limitations}

\input{figures_suppmat/ablation/ablation}

\input{figures_suppmat/ablation/ablation_additional}

\paragraph{Real-world evaluation.}
We extend the comparison with Neural Haircut~\cite{sklyarova2023neural_haircut} in Fig.~\ref{fig:geom_compare_suppmat}.
We also provide the reconstruction results for additional scenes in Fig.~\ref{fig:geom_compare_suppmat_new}-\ref{fig:short_hair_results}.
An extra comparison with the one-shot hair reconstruction method named HairStep~\cite{zheng2023hairstep} is provided in Fig.~\ref{fig:hairstep_comparison}.
We also expand the ablation study of the strands fitting in Figs~\ref{fig:ablation_compare_suppmat}-\ref{fig:ablation_compare_suppmat_additional}.
Besides the experiments presented in the main paper, we provide the close-ups and three more experiments: we remove color rendering loss $\mathcal{L}_\text{rgb}$, orientation rendering loss $\mathcal{L}_\text{dir}$ and diffusion-based loss $\mathcal{L}_\text{sds}$.
We observe that each of the components of our method meaningfully contribute to the final performance, as illustrated on the close-up images.

\paragraph{Limitations.}

The main limitation of our method follows Neural Haircut~\cite{sklyarova2023neural_haircut} and is related to the modeling of curly hairstyles, see Fig.~\ref{fig:limitations_suppmat}. 
We attribute this behavior to the root-to-tip design of the strand-based prior.
In our work, we use the same variant of the prior as the one proposed in Neural Haircut.
Improving the performance for the curly hairstyles thus remains future work.

%% file: figures_suppmat/comparison/comparison.tex
\begin{figure*}
    \includegraphics[width=\linewidth]{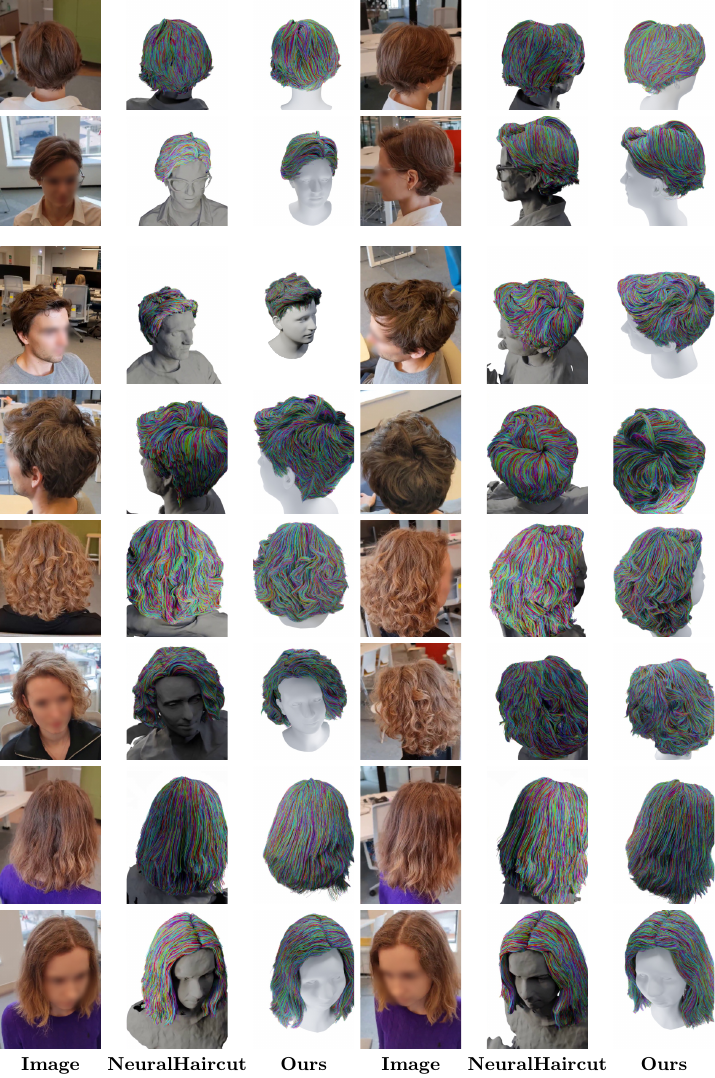}
    \caption{Additional qualitative comparison of the reconstructed strand-based geometry against Neural Haircut~\cite{sklyarova2023neural_haircut}. Slight misalignment appears due to test-view cameras interpolation.}
    \label{fig:geom_compare_suppmat}
\end{figure*}

%% file: figures_suppmat/comparison_2/comparison_new.tex
\begin{figure}[!t]
    \includegraphics[width=\linewidth]{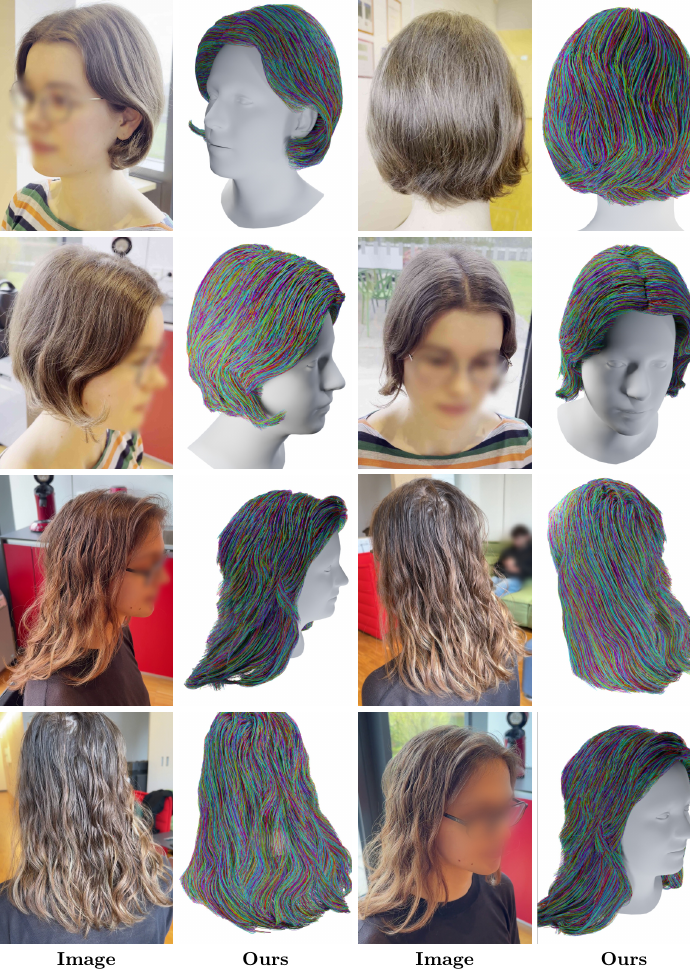}
    \caption{Qualitative results of our method on additional scenes. Slight misalignment appears due to test-view cameras interpolation.}
    \label{fig:geom_compare_suppmat_new}
\end{figure}

%% file: figures_suppmat/comparison_2/comparison2_2.tex
\begin{figure}[!t]
        

    \includegraphics[width=\linewidth]{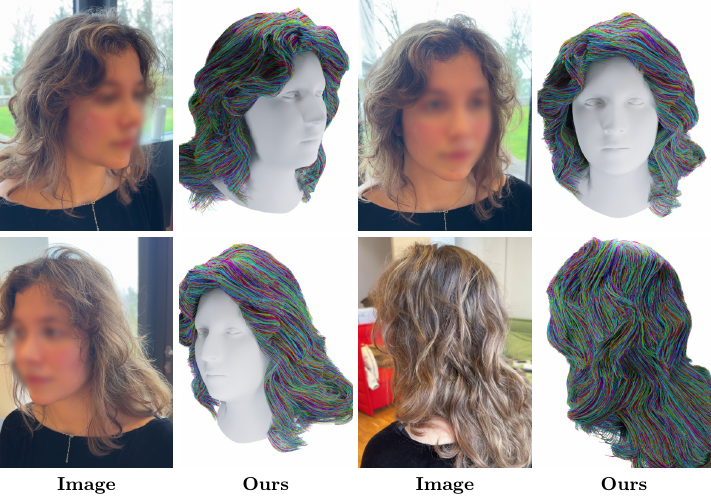}
    \caption{Qualitative results of our method on additional scenes. Slight misalignment appears due to test-view cameras interpolation.}
    \label{fig:geom_compare_suppmat_new2}
\end{figure}

%% file: figures_suppmat/short_hair.tex
\begin{figure}[!t]
    \begin{center}
    \includegraphics[width=\linewidth]{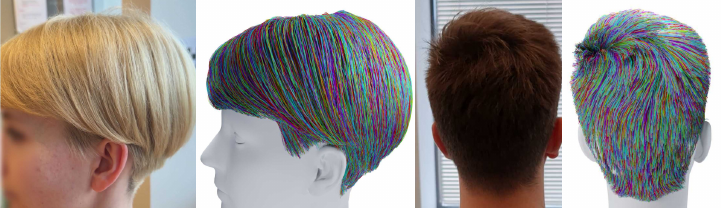}
    \caption{
        Qualitative results of our method on additional scenes. Slight misalignment appears due to test-view cameras interpolation. 
    } 
    \label{fig:short_hair_results}
    \end{center}
\end{figure}

%% file: figures_suppmat/hairstep_comparison.tex
\begin{figure}[!t]
    \begin{center}
    \includegraphics[width=\linewidth]{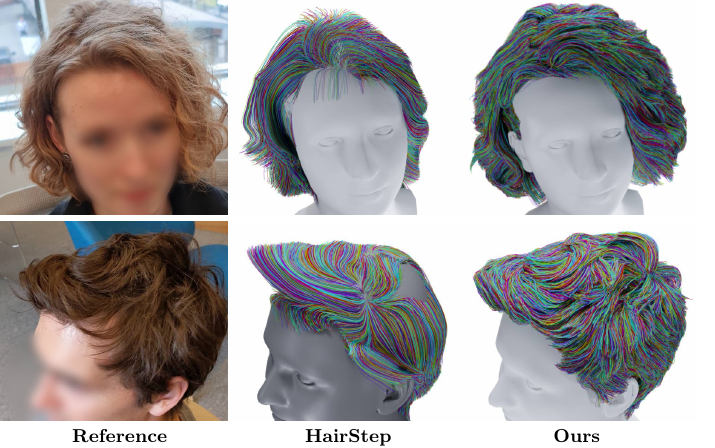}
    \caption{
        Comparison with the HairStep one-shot hair reconstruction method. We use a frontal image to infer the HairStep result.
    } 
    \label{fig:hairstep_comparison}
    \end{center}
\end{figure}

%% file: figures_suppmat/limitations/limitations.tex
\begin{figure*}[!t]
        
    \includegraphics[width=\linewidth]{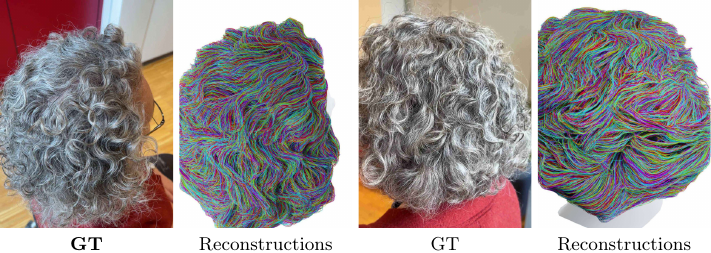}
    \caption{The limitations of our method include the problem of curly hairstyles modeling.}
    \label{fig:limitations_suppmat}
\end{figure*}

%% file: figures_suppmat/ablation/ablation.tex
\begin{figure*}[!t]
    \includegraphics[width=\linewidth]{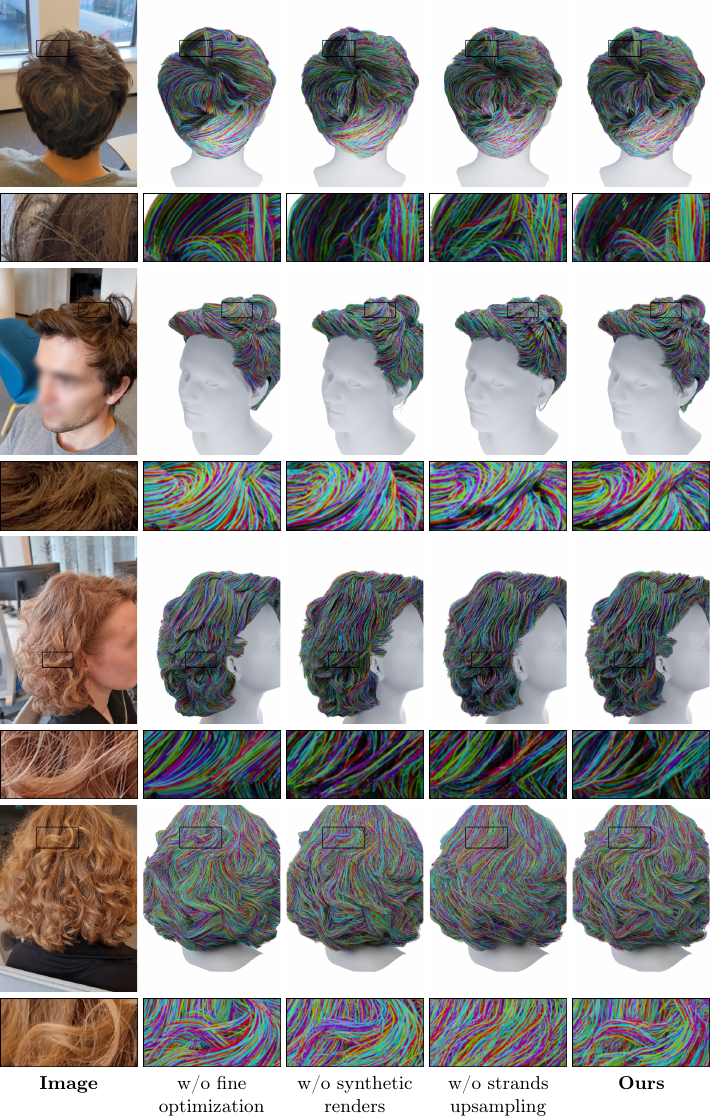}
    \caption{Extended ablation study of the strands fitting stage (Part 1).  
    }
    \label{fig:ablation_compare_suppmat}
\end{figure*}

%% file: figures_suppmat/ablation/ablation_additional.tex
\begin{figure*}[!t]
    \includegraphics[width=\linewidth]{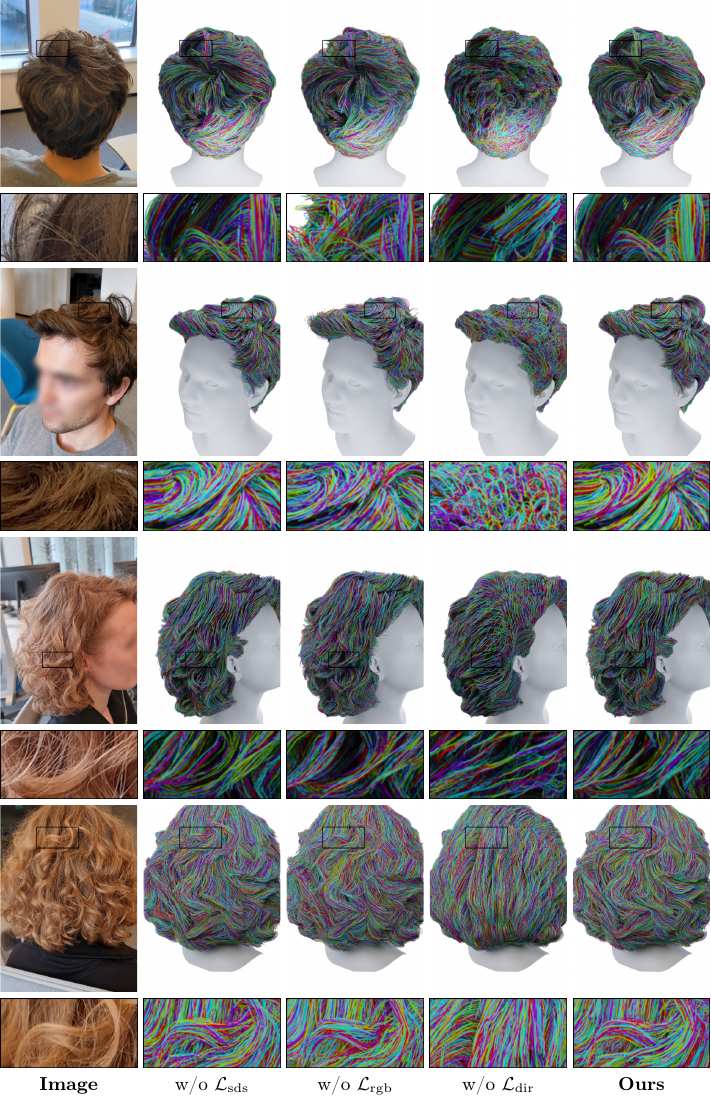}
    \caption{Extended ablation study of the strands fitting stage (Part 2). }
    \label{fig:ablation_compare_suppmat_additional}
\end{figure*}

%% file: arxiv.bbl
\begin{thebibliography}{10}
\providecommand{\url}[1]{\texttt{#1}}
\providecommand{\urlprefix}{URL }
\providecommand{\doi}[1]{https://doi.org/#1}

\bibitem{cao2013facewarehouse}
Cao, C., Weng, Y., Zhou, S., Tong, Y., Zhou, K.: Facewarehouse: A 3d facial expression database for visual computing. IEEE Transactions on Visualization and Computer Graphics  \textbf{20}(3),  413--425 (2013)

\bibitem{chai2015high}
Chai, M., Luo, L., Sunkavalli, K., Carr, N., Hadap, S., Zhou, K.: High-quality hair modeling from a single portrait photo. ACM Transactions on Graphics  \textbf{34}(6),  1--10 (2015)

\bibitem{Chai2016AutoHairFA}
Chai, M., Shao, T., Wu, H., Weng, Y., Zhou, K.: Autohair: fully automatic hair modeling from a single image. ACM Trans. Graph.  \textbf{35},  116:1--116:12 (2016)

\bibitem{chen2023monogaussianavatar}
Chen, Y., Wang, L., Li, Q., Xiao, H., Zhang, S., Yao, H., Liu, Y.: Monogaussianavatar: Monocular gaussian point-based head avatar. arXiv preprint arXiv:2312.04558  (2023)

\bibitem{chiang2016practical}
Chiang, M.J.Y., Bitterli, B., Tappan, C., Burley, B.: A practical and controllable hair and fur model for production path tracing. In: Computer Graphics Forum. vol.~35, pp. 275--283. Wiley Online Library (2016)

\bibitem{Blender}
Community, B.O.: Blender - a 3D modelling and rendering package. Blender Foundation, Stichting Blender Foundation, Amsterdam (2023), \url{http://www.blender.org}

\bibitem{daviet2023interactive}
Daviet, G.: Interactive hair simulation on the gpu using admm. In: ACM SIGGRAPH 2023 Conference Proceedings. pp. 1--11 (2023)

\bibitem{dhamo2023headgas}
Dhamo, H., Nie, Y., Moreau, A., Song, J., Shaw, R., Zhou, Y., P{\'e}rez-Pellitero, E.: Headgas: Real-time animatable head avatars via 3d gaussian splatting. arXiv preprint arXiv:2312.02902  (2023)

\bibitem{unrealengine}
{Epic Games}: Unreal engine, \url{https://www.unrealengine.com}

\bibitem{fascione2018path}
Fascione, L., Hanika, J., Piek{\'e}, R., Villemin, R., Hery, C., Gamito, M., Emrose, L., Mazzone, A.: Path tracing in production. In: ACM SIGGRAPH 2018 Courses, pp. 1--79 (2018)

\bibitem{fei2017multi}
Fei, Y., Maia, H.T., Batty, C., Zheng, C., Grinspun, E.: A multi-scale model for simulating liquid-hair interactions. ACM Transactions on Graphics (TOG)  \textbf{36}(4),  1--17 (2017)

\bibitem{garbin2022voltemorph}
Garbin, S.J., Kowalski, M., Estellers, V., Szymanowicz, S., Rezaeifar, S., Shen, J., Johnson, M., Valentin, J.: Voltemorph: Realtime, controllable and generalisable animation of volumetric representations. arXiv preprint arXiv:2208.00949  (2022)

\bibitem{hsu2023sagfree}
Hsu, J., Wang, T., Pan, Z., Gao, X., Yuksel, C., Wu, K.: Sag-free initialization for strand-based hybrid hair simulation. ACM Transactions on Graphics (Proceedings of SIGGRAPH 2023)  \textbf{42}(4) (07 2023)

\bibitem{Jiang2023HiFi4GHH}
Jiang, Y., Shen, Z., Wang, P., Su, Z., Hong, Y., Zhang, Y., Yu, J., Xu, L.: Hifi4g: High-fidelity human performance rendering via compact gaussian splatting. ArXiv  \textbf{abs/2312.03461} (2023)

\bibitem{Karras2022ElucidatingTD}
Karras, T., Aittala, M., Aila, T., Laine, S.: Elucidating the design space of diffusion-based generative models. In: Advances in Neural Information Processing Systems (NeurIPS) (2022)

\bibitem{kerbl3Dgaussians}
Kerbl, B., Kopanas, G., Leimk{\"u}hler, T., Drettakis, G.: 3d gaussian splatting for real-time radiance field rendering. ACM Transactions on Graphics  \textbf{42}(4) (July 2023)

\bibitem{kirschstein2023nersemble}
Kirschstein, T., Qian, S., Giebenhain, S., Walter, T., Nie\ss{}ner, M.: Nersemble: Multi-view radiance field reconstruction of human heads. ACM Trans. Graph.  \textbf{42}(4) (jul 2023)

\bibitem{FLAME:SiggraphAsia2017}
Li, T., Bolkart, T., Black, M.J., Li, H., Romero, J.: Learning a model of facial shape and expression from {4D} scans. ACM Transactions on Graphics, (Proc. SIGGRAPH Asia)  \textbf{36}(6),  194:1--194:17 (2017)

\bibitem{Lin2021BARFBN}
Lin, C.H., Ma, W.C., Torralba, A., Lucey, S.: Barf: Bundle-adjusting neural radiance fields. 2021 IEEE/CVF International Conference on Computer Vision (ICCV) pp. 5721--5731 (2021)

\bibitem{lombardi2021mixture}
Lombardi, S., Simon, T., Schwartz, G., Zollhoefer, M., Sheikh, Y., Saragih, J.: Mixture of volumetric primitives for efficient neural rendering. ACM Transactions on Graphics (ToG)  \textbf{40}(4),  1--13 (2021)

\bibitem{Lombardi2021MixtureOV}
Lombardi, S., Simon, T., Schwartz, G., Zollhoefer, M., Sheikh, Y., Saragih, J.M.: Mixture of volumetric primitives for efficient neural rendering. ACM Transactions on Graphics (TOG)  \textbf{40},  1 -- 13 (2021)

\bibitem{Luo2024GaussianHairHM}
Luo, H., Min, O., Zhao, Z., Jiang, S., Zhang, L., Zhang, Q., Yang, W., Xu, L., Yu, J.: Gaussianhair: Hair modeling and rendering with light-aware gaussians. vol. abs/2402.10483 (2024)

\bibitem{luo2012multi}
Luo, L., Li, H., Paris, S., Weise, T., Pauly, M., Rusinkiewicz, S.: Multi-view hair capture using orientation fields. In: 2012 IEEE Conference on Computer Vision and Pattern Recognition. pp. 1490--1497. IEEE (2012)

\bibitem{luo2013structure}
Luo, L., Li, H., Rusinkiewicz, S.: Structure-aware hair capture. ACM Transactions on Graphics  \textbf{32}(4),  1--12 (2013)

\bibitem{luo2013wide}
Luo, L., Zhang, C., Zhang, Z., Rusinkiewicz, S.: Wide-baseline hair capture using strand-based refinement. In: Proceedings of the IEEE Conference on Computer Vision and Pattern Recognition. pp. 265--272 (2013)

\bibitem{Nam2019StrandAccurateMH}
Nam, G., Wu, C., Kim, M.H., Sheikh, Y.: Strand-accurate multi-view hair capture. 2019 IEEE/CVF Conference on Computer Vision and Pattern Recognition (CVPR) pp. 155--164 (2019)

\bibitem{Paris2004CaptureOH}
Paris, S., Brice{\~n}o, H.M., Sillion, F.X.: Capture of hair geometry from multiple images. ACM SIGGRAPH 2004 Papers  (2004)

\bibitem{paris2008hair}
Paris, S., Chang, W., Kozhushnyan, O.I., Jarosz, W., Matusik, W., Zwicker, M., Durand, F.: Hair photobooth: geometric and photometric acquisition of real hairstyles. ACM Transactions on Graphics  \textbf{27}(3), ~30 (2008)

\bibitem{piuze2011generalized}
Piuze, E., Kry, P.G., Siddiqi, K.: Generalized helicoids for modeling hair geometry. In: Computer Graphics Forum. vol.~30, pp. 247--256. Wiley Online Library (2011)

\bibitem{Poole2022DreamFusionTU}
Poole, B., Jain, A., Barron, J.T., Mildenhall, B.: Dreamfusion: Text-to-3d using 2d diffusion. ICLR  \textbf{abs/2209.14988} (2023)

\bibitem{qian2023gaussianavatars}
Qian, S., Kirschstein, T., Schoneveld, L., Davoli, D., Giebenhain, S., Nie\ss{}ner, M.: Gaussianavatars: Photorealistic head avatars with rigged 3d gaussians. arXiv preprint arXiv:2312.02069  (2023)

\bibitem{rivero2024rig3dgs}
Rivero, A., Athar, S., Shu, Z., Samaras, D.: Rig3dgs: Creating controllable portraits from casual monocular videos. arXiv preprint arXiv:2402.03723  (2024)

\bibitem{Rosu2022NeuralSL}
Rosu, R.A., Saito, S., Wang, Z., Wu, C., Behnke, S., Nam, G.: Neural strands: Learning hair geometry and appearance from multi-view images. In: European Conference on Computer Vision (2022)

\bibitem{Saito2023RelightableGC}
Saito, S., Schwartz, G., Simon, T., Li, J., Nam, G.: Relightable gaussian codec avatars. ArXiv  \textbf{abs/2312.03704} (2023)

\bibitem{schoenberger2016sfm}
Sch\"{o}nberger, J.L., Frahm, J.M.: Structure-from-motion revisited. In: Conference on Computer Vision and Pattern Recognition (CVPR) (2016)

\bibitem{shen2023CT2Hair}
Shen, Y., Saito, S., Wang, Z., Maury, O., Wu, C., Hodgins, J., Zheng, Y., Nam, G.: Ct2hair: High-fidelity 3d hair modeling using computed tomography. ACM Transactions on Graphics  \textbf{42}(4),  1--13 (2023)

\bibitem{shen2020deepsketchhair}
Shen, Y., Zhang, C., Fu, H., Zhou, K., Zheng, Y.: Deepsketchhair: Deep sketch-based 3d hair modeling. IEEE transactions on visualization and computer graphics  \textbf{27}(7),  3250--3263 (2020)

\bibitem{sklyarova2023neural_haircut}
Sklyarova, V., Chelishev, J., Dogaru, A., Medvedev, I., Lempitsky, V., Zakharov, E.: Neural haircut: Prior-guided strand-based hair reconstruction. In: Proceedings of IEEE International Conference on Computer Vision (ICCV) (2023)

\bibitem{sklyarova2023haar}
Sklyarova, V., Zakharov, E., Hilliges, O., Black, M.J., Thies, J.: Haar: Text-conditioned generative model of 3d strand-based human hairstyles. ArXiv  (Dec 2023)

\bibitem{IQA}
Su, S., Yan, Q., Zhu, Y., Zhang, C., Ge, X., Sun, J., Zhang, Y.: Blindly assess image quality in the wild guided by a self-adaptive hyper network. 2020 IEEE/CVF Conference on Computer Vision and Pattern Recognition (CVPR) pp. 3664--3673 (2020)

\bibitem{Wang2021NeuSLN}
Wang, P., Liu, L., Liu, Y., Theobalt, C., Komura, T., Wang, W.: Neus: Learning neural implicit surfaces by volume rendering for multi-view reconstruction. In: Advances in Neural Information Processing Systems (NeurIPS) (2022)

\bibitem{Wang2022NeuWigsAN}
Wang, Z., Nam, G., Stuyck, T., Lombardi, S., Cao, C., Saragih, J.M., Zollhoefer, M., Hodgins, J.K., Lassner, C.: Neuwigs: A neural dynamic model for volumetric hair capture and animation. In: Proceedings of the IEEE/CVF Conference on Computer Vision and Pattern Recognition (CVPR). pp. 8641--8651 (June 2023)

\bibitem{Wang2021HVHLA}
Wang, Z., Nam, G., Stuyck, T., Lombardi, S., Zollhoefer, M., Hodgins, J.K., Lassner, C.: Hvh: Learning a hybrid neural volumetric representation for dynamic hair performance capture. 2022 IEEE/CVF Conference on Computer Vision and Pattern Recognition (CVPR) pp. 6133--6144 (2021)

\bibitem{woo1999opengl}
Woo, M., Neider, J., Davis, T., Shreiner, D.: OpenGL programming guide: the official guide to learning OpenGL, version 1.2. Addison-Wesley Longman Publishing Co., Inc. (1999)

\bibitem{xiang2023flashavatar}
Xiang, J., Gao, X., Guo, Y., Zhang, J.: Flashavatar: High-fidelity digital avatar rendering at 300fps (2023)

\bibitem{xing2019hairbrush}
Xing, J., Nagano, K., Chen, W., Xu, H., Wei, L.y., Zhao, Y., Lu, J., Kim, B., Li, H.: Hairbrush for immersive data-driven hair modeling. In: Proceedings of the 32Nd Annual ACM Symposium on User Interface Software and Technology. pp. 263--279 (2019)

\bibitem{xu2023gaussianheadavatar}
Xu, Y., Chen, B., Li, Z., Zhang, H., Wang, L., Zheng, Z., Liu, Y.: Gaussian head avatar: Ultra high-fidelity head avatar via dynamic gaussians  (2023)

\bibitem{matte_anything}
Yao, J., Wang, X., Ye, L., Liu, W.: Matte anything: Interactive natural image matting with segment anything models. arXiv preprint arXiv:2306.04121  (2023)

\bibitem{Yuksel2009HairM}
Yuksel, C., Schaefer, S., Keyser, J.: Hair meshes. ACM Transactions on Graphics  \textbf{28}(5), ~1--7 (2009)

\bibitem{zhang2017data}
Zhang, M., Chai, M., Wu, H., Yang, H., Zhou, K.: A data-driven approach to four-view image-based hair modeling. ACM Transactions on Graphics  \textbf{36}(4),  156--1 (2017)

\bibitem{zhang2018modeling}
Zhang, M., Wu, P., Wu, H., Weng, Y., Zheng, Y., Zhou, K.: Modeling hair from an rgb-d camera. ACM Transactions on Graphics  \textbf{37}(6),  1--10 (2018)

\bibitem{zhao2024psavatar}
Zhao, Z., Bao, Z., Li, Q., Qiu, G., Liu, K.: Psavatar: A point-based morphable shape model for real-time head avatar creation with 3d gaussian splatting. arXiv preprint arXiv:2401.12900  (2024)

\bibitem{zheng2023hairstep}
Zheng, Y., Jin, Z., Li, M., Huang, H., Ma, C., Cui, S., Han, X.: Hairstep: Transfer synthetic to real using strand and depth maps for single-view 3d hair modeling. In: Proceedings of the IEEE/CVF Conference on Computer Vision and Pattern Recognition. pp. 12726--12735 (2023)

\bibitem{zhou2023groomgen}
Zhou, Y., Chai, M., Pepe, A., Gross, M., Beeler, T.: Groomgen: A high-quality generative hair model using hierarchical latent representations. ACM Transactions on Graphics (TOG)  \textbf{42}(6),  1--16 (2023)

\bibitem{zielonka2023instant}
Zielonka, W., Bolkart, T., Thies, J.: Instant volumetric head avatars. In: Proceedings of the IEEE/CVF Conference on Computer Vision and Pattern Recognition. pp. 4574--4584 (2023)

\end{thebibliography}
